\documentclass[letterpaper, 10 pt, conference]{ieeeconf}

\usepackage{times}
\usepackage{amsmath,amssymb}
\usepackage{multicol}
\usepackage{graphicx}
\usepackage{booktabs}
\usepackage{multirow}
\usepackage{threeparttable}
\usepackage[compatibility=false]{caption}
\usepackage{subcaption}
\usepackage{flushend}
\usepackage{xcolor}
\usepackage{algorithm}
\usepackage{algpseudocode}
\usepackage{cite}
\usepackage{microtype}

\IEEEoverridecommandlockouts                              % This command is only needed if 
\title{\LARGE \bf
Uncertainty-Driven Self-Play for Robust Multi-Agent Driving Policies
}

\author{Sajad Marvi$^{1,2}$, Julian Schmidt$^{1}$, Alessandro Canevaro$^{1}$, Julian Jordan$^{1}$, and Abhinav Valada$^2$%  <-this % stops a space
\thanks{$^{1}$Mercedes-Benz AG, Germany.}%
\thanks{$^{2}$Department of Computer Science, University of Freiburg, Germany.}%
}
\begin{document}

\maketitle
\thispagestyle{empty}
\pagestyle{empty}

\begin{abstract}
Learning robust driving policies that generalize beyond their training domain remains a central challenge in motion planning. We propose an uncertainty-driven self-play scheme, in which half of the agents are rewarded for \emph{increasing} their neighbors' epistemic uncertainty, while the other half learn to drive safely as the exploring agents continuously seek out their weaknesses. Epistemic uncertainty indicates where the policy is least confident and serves as an automatic proxy for scenario difficulty. Exploring agents use this signal to seek out hard cases, generating a challenging curriculum without manual scenario design. The resulting policy outperforms standard self-play, achieving best-in-class zero-shot transfer to nuplan Val14 (99.12\% vs.\ 97.97\%) and Argoverse~2 (82.39\% vs.\ 78.03\%) without fine-tuning, while also improving performance over standard self-play on Waymo. Our diversity-driven approach achieves 85\% higher LSTM hidden-state coverage than the baseline, indicating a much richer set of behaviors. Further analysis shows that epistemic uncertainty serves as a calibrated difficulty detector, with highest-uncertain agents exhibiting 14.8$\times$ the combined failure rate across collision, off-road, and DNF, of lowest-uncertain agents, enabling automatic prioritization of challenging scenarios without ground-truth labels.
\end{abstract}

\section{Introduction}

Object-level behavior planning is a core challenge in autonomous driving where a policy must control agents through complex multi-agent scenarios involving dynamic interactions with other vehicles, pedestrians, and road structure. Imitation learning~\cite{caesar2020nuscenesmultimodaldatasetautonomous,wilson2023argoverse2generativedatasets,schmidt2025graphpilot} trains policies from logged human trajectories but underrepresents rare, safety-critical events. Reinforcement learning~\cite{cornelisse2025building,wei2026copark,wei2026counterplay, schmalstieg2023learning,schmalstieg2022learning} scales well and produces competent drivers, but policies reflect only the scenarios seen during training and may not generalize to novel domains. Training separate adversarial policies~\cite{wang2021advsim,hanselmann2022kinggeneratingsafetycriticaldriving} can expose the ego agent to harder situations, but such policies are expensive to maintain and tend to produce unrealistic behaviors. 
A promising alternative is conditioned self-play, where all agents share one network and a conditioning signal selects the behavioral mode for each agent. Prior work conditions on return targets~\cite{rowe2024ctrlsim}, skill indices~\cite{eysenbach2019diayn}, or entropy bonuses~\cite{haarnoja2018sac}, but these face a fundamental tension. Achieving meaningful behavioral diversity typically degrades driving performance, even when conditioned to drive normally. We term this the \emph{diversity tax}. Can a conditioned policy, set to its normal driving mode, not only match but \emph{exceed} an unconditional self-play baseline?

Our key insight is that epistemic uncertainty, which measures the policy's confidence, provides a natural proxy for behavioral diversity. We train half the agents to \emph{increase} their neighbors' uncertainty while the other half learn to drive safely under this probing pressure. Uncertainty is an inexpensive, automatic proxy for scenario difficulty. Rather than hand-engineering diverse training scenarios, probing agents find the hard cases automatically by targeting exactly where the current policy is most confused. This self-play scheme generates an automatic diversity-driven curriculum of challenging interactions. Among the conditioned methods we evaluate, our method is the only one to outperform a strong unconditional self-play baseline.

\begin{figure}[t]
    \centering
    \includegraphics[width=\columnwidth]{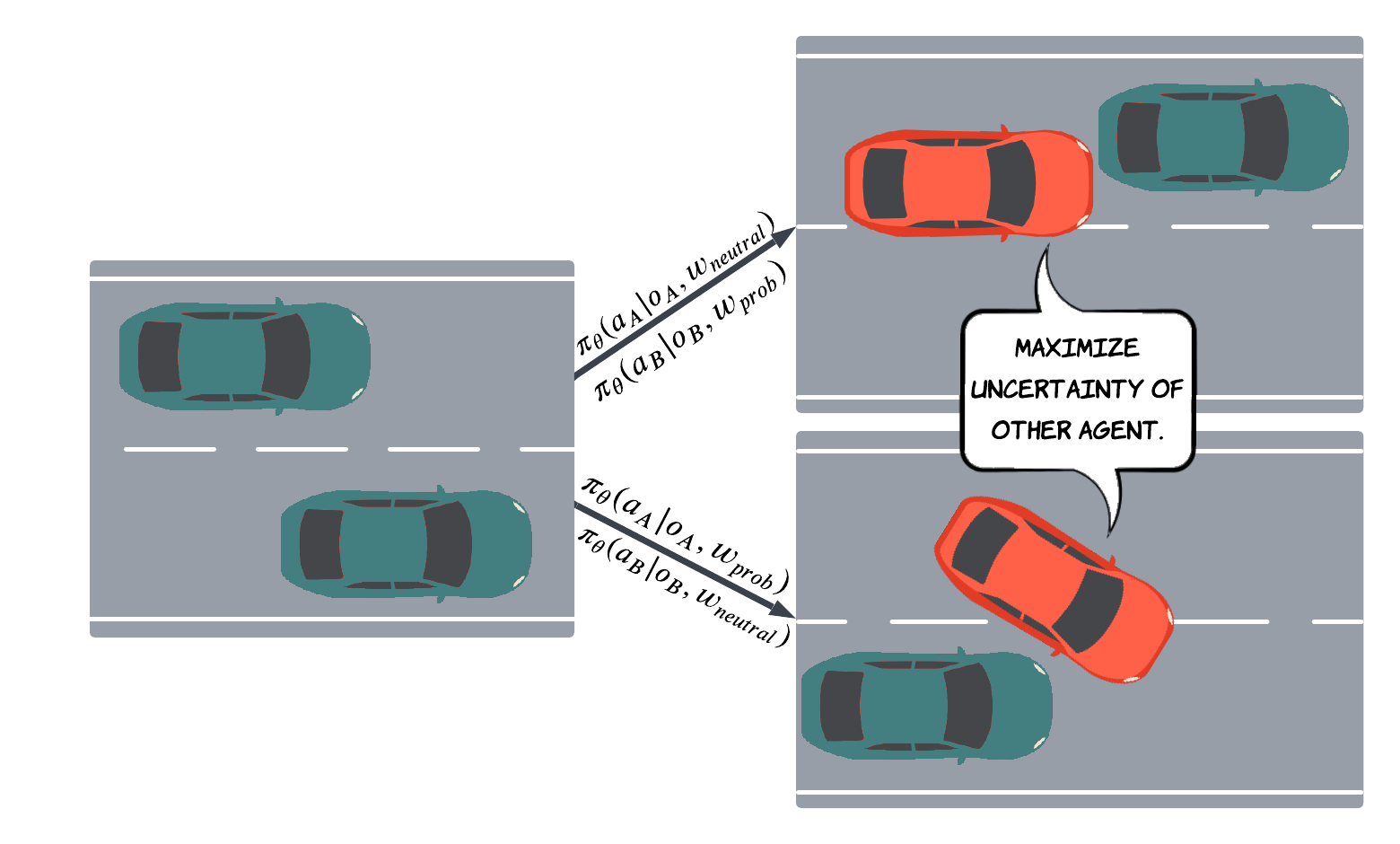}
    \caption{Probing agents (red), assigned the role in each episode, are rewarded for increasing the epistemic uncertainty of their nearest neighbors, while neutral agents (blue) receive only the task reward. As neutral agents adapt, probing agents must discover increasingly novel confusing situations, generating an automatic diversity-driven curriculum. At evaluation, all agents use the neutral role, benefiting from the broader training distribution.}
    \label{fig:intro}
    \vspace{-0.2cm}
\end{figure}

In each training episode (Fig.~\ref{fig:intro}), 50\% of agents are designated \emph{probing agents} (rewarded for increasing the epistemic uncertainty of their $K{=}3$ nearest neighbors) while 50\% are \emph{neutral} (pure task reward). As neutral agents adapt, probing agents must discover increasingly novel challenging situations, generating a diversity-driven curriculum without hand-crafted scenario design. At evaluation, all agents use the neutral role, benefiting from exposure to this broader training distribution. This relates to adversarial training~\cite{madry2018towards} and domain randomization~\cite{tobin2017domain} where both improve robustness by exposing the learner to challenging conditions during training.

Our contributions are:
\begin{enumerate}
    \item An uncertainty-driven self-play scheme using epistemic uncertainty change ($\Delta$unc) as the probing reward, showing that a conditioned policy can surpass an unconditional self-play baseline, an outcome not achieved by any other conditioned method we evaluate.
    \item Empirical demonstration that epistemic uncertainty serves as a calibrated difficulty detector, with highest-uncertainty scenarios exhibiting 14.8$\times$ the combined failure rate across collision, off-road, and DNF, of lowest-uncertainty scenarios (ECE$=$0.014).
    \item Comprehensive evaluations across 10,000 Waymo scenarios and zero-shot transfer to nuplan Val14 and Argoverse~2, with comparisons against DIAYN, entropy-conditioned, and GIGAFLOW-style baselines.
    \item Ablation studies identifying neighborhood size ($K{=}3$), warmup scheduling, conditioning mechanism (FiLM vs.\ Concatenation vs.\ none), and probing fraction (50\%) as essential design choices, with warmup improving score by 0.57\% and role conditioning being load-bearing for any gains over the unconditional baseline.

\end{enumerate}

The resulting policy outperforms standard self-play, achieving best-in-class zero-shot transfer to nuplan Val14 (99.12\% vs.\ 97.97\%) and Argoverse~2 (82.39\% vs.\ 78.03\%) with no fine-tuning, while reducing the failure rate by 17\% on Waymo validation scenarios compared to the baseline. It also achieves 85\% higher LSTM hidden-state coverage, indicating a broader range of learned behaviors that explains the improved transfer.

\section{Related Work}
Our work spans controllable traffic simulation, adversarial training, diversity-promoting RL, and uncertainty estimation.

\noindent\textbf{Diverse Traffic Scenario Generation}: 
Generating diverse traffic behavior is critical for Autonomous Vehicle~(AV) validation. CtRL-Sim~\cite{rowe2024ctrlsim} conditions a diffusion model on return targets to yield reactive agents of varying aggressiveness, but requires offline trajectory data and separate generative models. MATS-Gym~\cite{matsgym2024} provides a framework for multi-agent traffic scenarios but does not address uncertainty-driven curriculum generation. GIGAFLOW~\cite{cusumano2025robust} demonstrates that massive-scale self-play with reward-weight conditioning produces robust driving with some behavioral variation, but conditions on task reward parameters (e.g., collision penalty magnitude), which couples diversity to driving competence. In contrast, our method (UncArms) conditions only on a scalar role signal derived from epistemic uncertainty, keeping task reward weights fixed and avoiding the diversity tax entirely.

\noindent\textbf{Adversarial Training for Robustness}: 
Adversarial training improves model robustness by exposing learners to worst-case perturbations~\cite{madry2018towards}. In robotics, domain randomization~\cite{tobin2017domain} and adversarial environment generation~\cite{dennis2020emergent} create challenging training conditions that improve transfer. For driving, RARL~\cite{pinto2017robust} trains a protagonist against an adversary, but uses fixed adversarial objectives rather than adaptive ones. Dedicated adversarial scenario generators such as SaFeR~\cite{safer2026} and IntSim~\cite{intsim2025} optimize for safety-critical situations but require separate training loops or privileged world-model access. Our approach differs in using the \emph{learner's own epistemic uncertainty} within a single shared policy, creating an adaptive curriculum without separate adversarial optimization.

\noindent\textbf{Diversity in Reinforcement Learning}: 
DIAYN~\cite{eysenbach2019diayn} discovers diverse skills by maximizing mutual information between skill indices and states, but the discrete skill space limits continuous behavioral interpolation and does not target policy weaknesses. Maximum entropy RL~\cite{haarnoja2018sac} encourages exploration through entropy bonuses but does not target specific policy weaknesses. Population-based approaches~\cite{vinyals2019grandmaster} maintain diversity through separate agents, incurring proportional computational costs. Our approach targets \emph{specific} weaknesses (where the policy is uncertain) rather than undirected diversity.

\noindent\textbf{Epistemic Uncertainty in Decision-Making}: 
Monte Carlo Dropout~\cite{gal2016dropoutbayesianapproximationrepresenting} provides computationally tractable epistemic uncertainty estimates. Prior work has used uncertainty for exploration bonuses in single-agent settings~\cite{pathak2017curiositydrivenexplorationselfsupervisedprediction,burda2018largescalestudycuriositydrivenlearning} and for safe navigation under uncertainty~\cite{lutjens2019saferl,kahn2017uncertaintyaware}. Our key insight is that uncertainty can serve simultaneously as a calibrated difficulty detector (14.8$\times$ combined failure rate across deciles) and as a diversity-driving training signal in multi-agent settings.

\begin{figure*}[t]
    \centering
    \includegraphics[width=0.7\textwidth]{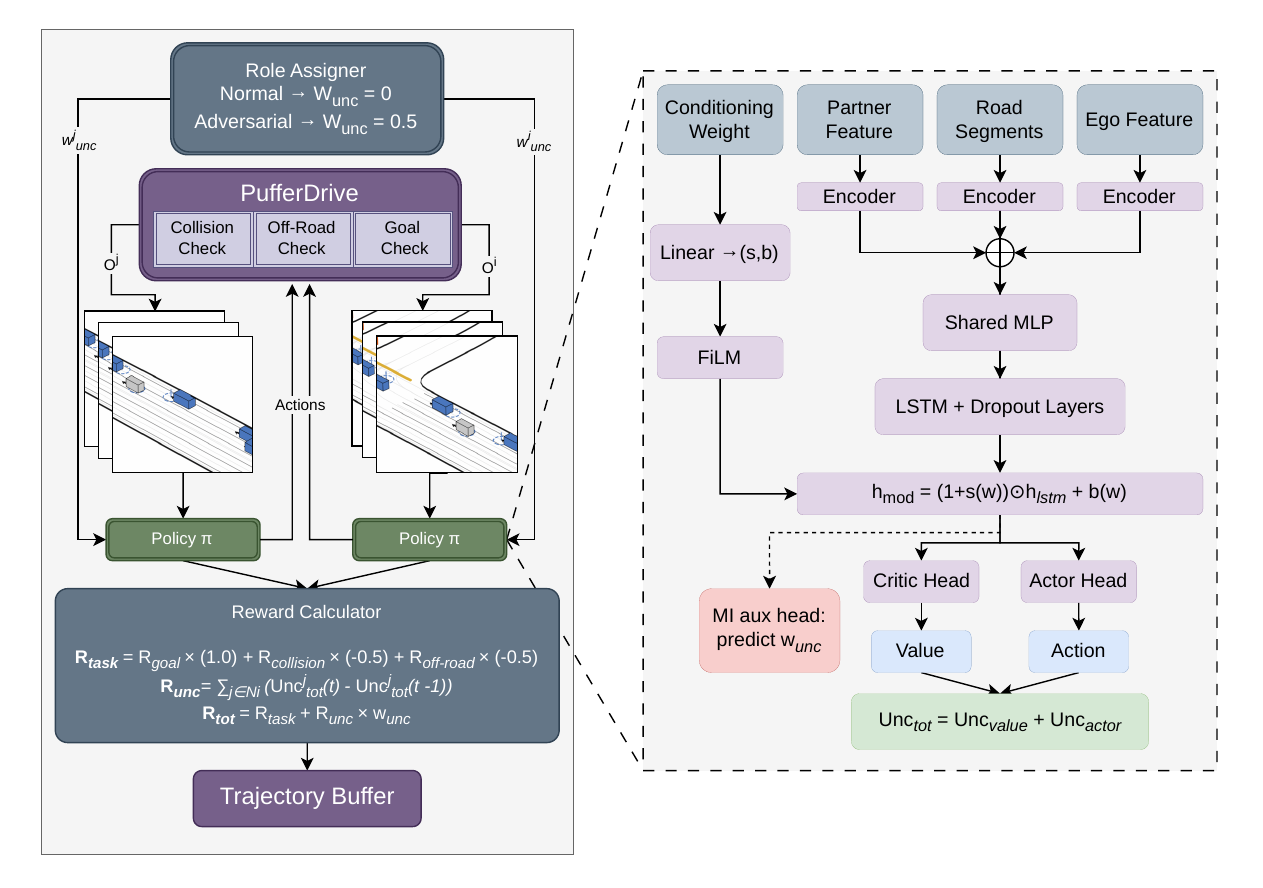}
    \caption{Architecture overview. Observations are processed by ego, partner, and road encoders with max-pooling over variable-count entities. The concatenated embedding passes through an LSTM with MC-Dropout and is modulated by FiLM conditioning on a scalar role signal (probing or neutral). At evaluation, all agents use the neutral role.}
    \label{fig:arc}
    \vspace{-0.2cm}
\end{figure*}

\section{Method}
Our method, UncArms (Uncertainty Arms-Race Self-Play), trains a shared policy through uncertainty-driven self-play in which agents are assigned one of two roles per episode: probing or neutral. We use \emph{arms race} in the co-evolutionary sense. As neutral agents adapt, probing agents must escalate to maintain their reward, producing an escalating curriculum analogous to the Red Queen effect. This is not a zero-sum conflict, since neutral agents are never explicitly rewarded for resisting probing, but the adaptation pressure is nonetheless bidirectional. Both roles share the same FiLM-conditioned architecture. A scalar role signal allows the policy to specialize its behavior per role while sharing all parameters.

% Our method, UncArms (Uncertainty Arms-Race Self-Play), trains a shared policy through uncertainty-driven self-play in which agents are assigned one of two roles per episode: probing or neutral. Both roles share the same FiLM-conditioned architecture. A scalar role signal is passed as input, allowing the policy to specialize its behavior per role while sharing all parameters.

\subsection{Problem Formulation}

We formulate multi-agent driving as a Partially Observable Markov Game:
\begin{equation}
    (\mathcal{N}, \mathcal{S}, \{\mathcal{A}_i\}_{i \in \mathcal{N}}, \mathcal{P}, \{R_i\}_{i \in \mathcal{N}}, \Omega, \{\mathcal{O}_i\}_{i \in \mathcal{N}}, \gamma),
\end{equation}
where $\mathcal{N}$ is the set of $N{=}32$ agents per scenario, $\mathcal{S}$ the global state space, $\mathcal{P}$ the transition function, $R_i$ the reward function for agent $i$, $\Omega$ the joint observation space, and $\gamma{=}0.99$ the discount factor. Each agent's partial observation $\mathcal{O}_i$ consists of ego features (speed, heading, relative goal position, lane alignment), relative partner features (position, heading, speed) for nearby agents, road geometry (segment positions, orientations, lane types), and the conditioning weight $w_\mathrm{unc}$. Each agent selects actions from a discrete space $\mathcal{A}_i$ of 91 combinations (7 acceleration $\times$ 13 steering bins).

\subsection{Epistemic Uncertainty via MC-Dropout}

We estimate epistemic uncertainty using Monte Carlo Dropout~\cite{gal2016dropoutbayesianapproximationrepresenting} with $K{=}5$ stochastic forward passes during training rollouts. Critic uncertainty is the standard deviation across value estimates $V_k$ from the $k$-th forward pass, with $\bar{V}$ their mean as
\begin{equation}
    \mathrm{Unc}_\mathrm{value}^\mathrm{ep} = \sqrt{\frac{1}{K}\sum_{k=1}^{K}(V_k - \bar{V})^2}.
\end{equation}
Actor uncertainty uses mutual information decomposition, where $\pi_k(a|s)$ is the action distribution from the $k$-th forward pass and $H[\cdot]$ denotes Shannon entropy:
\begin{equation}
    \mathrm{Unc}_\mathrm{actor}^\mathrm{ep} = H\!\left[\frac{1}{K}\sum_{k=1}^{K}\pi_k(a|s)\right] - \frac{1}{K}\sum_{k=1}^{K}H[\pi_k(a|s)].
\end{equation}
The total uncertainty combines both via $\mathrm{Unc}_\mathrm{total} = \mathrm{Unc}_\mathrm{value}^\mathrm{ep} + \mathrm{Unc}_\mathrm{actor}^\mathrm{ep}$. Together, they capture complementary aspects of the policy's epistemic state in which the actor term reflects uncertainty about what action to take, while the critic term reflects uncertainty about how valuable the current situation is. This formulation assumes an actor-critic architecture, although the same concept applies to any method from which policy uncertainty can be estimated.
While this work focuses on the self-play mechanism rather than the uncertainty estimator itself, we chose MC-Dropout for its simplicity: $K{=}5$ stochastic passes through a single shared model, with no additional networks. Exploring alternative estimators such as deep ensembles~\cite{lakshminarayanan2017simple} or evidential methods~\cite{marvi2025evidential} is a natural extension that could reduce training overhead without changing the core arms-race dynamic.

\textbf{Important distinction:} Uncertainty computation occurs only during \emph{training rollouts} to generate rewards. At evaluation, all agents operate with neutral conditioning ($w_\mathrm{unc}{=}0$), requiring no uncertainty computation. The policy runs as a standard single-pass network with zero additional inference cost. Accessing neighbors' uncertainty is a \emph{training curriculum mechanism}, not a privileged observation at deployment.

\subsection{FiLM-Conditioned Architecture}
\label{sec:film}

The architecture (Fig.~\ref{fig:arc}) processes three observation modalities through separate encoders:
\begin{itemize}
    \item \textit{Ego encoder}: ego features $\rightarrow$ 64 $\rightarrow$ 64 (LayerNorm + GELU + Dropout).
    \item \textit{Partner encoder}: per-partner features $\rightarrow$ 64 $\rightarrow$ 64, max-pooled over all visible partners.
    \item \textit{Road encoder}: per-segment features $\rightarrow$ 64 $\rightarrow$ 64, max-pooled over road segments.
\end{itemize}

The concatenated 192-dimensional representation passes through a shared embedding layer (192 $\rightarrow$ 256, GELU) and an LSTM with Dropout Layer ($p{=}0.2$), then undergoes Feature-wise Linear Modulation (FiLM)~\cite{perez2018film}:
\begin{equation}
    \mathbf{h}_\mathrm{mod} = (1 + \boldsymbol{s}(\mathbf{w})) \odot \mathbf{h}_\mathrm{LSTM} + \boldsymbol{b}(\mathbf{w}),
\end{equation}
where $\boldsymbol{s}, \boldsymbol{b}$ are produced by a linear projection of the scalar role signal $w_\mathrm{role}$. Max-pooling over partner and road encodings naturally handles variable agent counts without padding or masking. The full model has about 780K parameters.

\textbf{Mutual information auxiliary loss.} To ensure the policy attends to the role signal rather than ignoring it, an auxiliary head predicts the conditioning input from the shared representation:
\begin{equation}
    \mathcal{L}_\mathrm{MI} = \| \hat{w}_\mathrm{role} - w_\mathrm{role} \|_2^2,
\end{equation}
weighted by $\alpha_\mathrm{MI} = 0.1$ relative to the PPO objective. This maximizes a lower bound on $I(w_\mathrm{role}; \mathbf{h})$, ensuring the representation differentiates probing and neutral roles.

\subsection{Uncertainty-Driven Self-Play via Uncertainty Change}
\label{sec:arms_race}

UncArms uses epistemic uncertainty as a \emph{training signal} (Fig.~\ref{fig:arc}, left): probing agents are rewarded for increasing neighbors' uncertainty, exposing the neutral policy to a broader training distribution.

\textbf{Role assignment.} At each episode, agents are randomly assigned roles: 50\% \emph{probing} (role signal $w_\mathrm{role}$ sampled uniformly from $(0, 0.5]$) and 50\% \emph{neutral} (role signal $w_\mathrm{role} = 0$). Neutral agents receive only the standard task reward. Probing agents receive an additional reward based on the uncertainty \emph{change} they induce in neighbors:
\begin{equation}
    R_\mathrm{adv}^i = \frac{1}{K} \sum_{j \in \mathrm{KNN}_K(i)} \left( \mathrm{Unc}_j(t) - \mathrm{Unc}_j(t{-}1) \right),
    \label{eq:adv_reward}
\end{equation}
where $\mathrm{KNN}_K(i)$ denotes the $K{=}3$ nearest agents to agent $i$, and $\Delta\mathrm{Unc}_j = \mathrm{Unc}_j(t) - \mathrm{Unc}_j(t{-}1)$ is the temporal change in epistemic uncertainty.

\textbf{Self-play dynamics.} The $\Delta$unc reward creates co-evolutionary pressure as probing agents must find situations that \emph{increase} neighbors' confusion (not just exploit existing uncertainty). As neutral agents adapt, the probing reward diminishes, forcing probing agents to discover increasingly novel confusing situations. This generates an automatic diversity-driven curriculum without hand-crafted scenario.

% \textbf{Training schedule.} UncArms uses a three-phase curriculum within its 1.5B steps:
% \begin{itemize}
%     \item Steps 0--750M: driving-only warmup (no adversarial reward), establishing base competence.
%     \item Steps 750M--1B: linear ramp of probing reward magnitude.
%     \item Steps 1B--1.5B: full probing intensity.
% \end{itemize}
\textbf{Training schedule.} UncArms uses a three-phase curriculum in which steps 0--750M are a driving-only warmup (no adversarial reward), steps 750M--1B linearly ramp the probing reward, and steps 1B--1.5B run at full probing intensity.

\textbf{Why $\Delta$unc rather than absolute uncertainty?} Rewarding absolute uncertainty would incentivize probing agents to merely approach already-confused agents. Rewarding the \emph{change} requires probing agents to actively create new confusion, driving exploration of the state space where the policy has unexplored weaknesses.

\section{Experimental Evaluation}

We evaluate all methods on the Waymo Open Motion Dataset (WOMD)~\cite{ettinger2021largescaleinteractivemotion} and test zero-shot transfer to nuPlan~\cite{nuplan} and Argoverse~2~\cite{wilson2023argoverse2generativedatasets}. All experiments use 4 seeds, and we report inter-seed mean$\pm$std.

\subsection{Setup}

\textbf{Simulator.} We train in PufferDrive\cite{pufferdrive2025github}, built on PufferLib~\cite{suarez2024pufferlibmakingreinforcementlearning} achieving $>$1M environment simulation steps/second with $N{=}32$ agents across 300 parallel WOMD scenarios~\cite{ettinger2021largescaleinteractivemotion}.

\textbf{Training.} All experiments use PPO~\cite{schulman2017proximalpolicyoptimizationalgorithms} with batch size 131,072, learning rate $3{\times}10^{-4}$, clip ratio 0.2, entropy coefficient $10^{-4}$, and discount $\gamma{=}0.99$. All methods train for 1.5B steps, following the setup in~\cite{pufferdrive2025github}.

\textbf{Hardware.} Training is performed on a single NVIDIA T4 GPU. Including neural network inference and PPO updates, effective training throughput is $\sim$46K steps/second. Each 1.5B-step run completes in approximately 9 hours.

\textbf{Evaluation.} Primary WOMD evaluation uses 10,000 held-out maps with all 32 agents policy-controlled (fully reactive multi-agent). Transfer evaluation uses SDC-only mode (single ego policy-controlled, background agents replay logged trajectories) to isolate the ego policy's generalization. All conditioned variants are evaluated at neutral conditioning ($w_\mathrm{unc}{=}0$).

\subsection{Baselines and Variants}

\begin{itemize}
    \item \textit{Baseline (Standard PPO)}: Self-play without FiLM, MC-Dropout, or conditioning~\cite{cornelisse2025building}. The unconditional baseline.
    \item \textit{GIGAFLOW$^\ast$}: Baseline architecture extended with GIGAFLOW-style~\cite{cusumano2025robust} concatenation-based conditioning on the 3 task reward weights ($w_\mathrm{goal}, w_\mathrm{coll}, w_\mathrm{off}$). No uncertainty signal. Evaluates the cost of task-weight conditioning.
    \item \textit{DIAYN}: Skill discovery with 5 discrete skills~\cite{eysenbach2019diayn}, maximizing $I(z; s)$.
    \item \textit{Ent-PPO (Entropy-conditioned PPO)}: Conditioned on target entropy level, with the entropy coefficient varied per agent per episode.
    % \item \textbf{UncControl (Ours):} FiLM conditioned on scalar $w_\mathrm{unc} \in [-0.05, 0.05]$, fixed task weights, MC-Dropout, MI auxiliary loss.
    \item \textit{UncArms (Ours)}: FiLM-conditioned architecture (Sec.~\ref{sec:film}), role signal conditioning, arms-race training with $\Delta$unc reward (50\% probing / 50\% neutral).
\end{itemize}

\subsection{Main Results}

\begin{table*}[t]
\centering
\caption{Main performance on WOMD (10K validation maps, 4 seeds, all methods trained for 1.5B steps). Score = goal reached $\wedge$ no collision. Goal = fraction reaching goal. $\dagger$: $p<0.05$ vs.\ Baseline (Welch's $t$-test). \textit{UncArms is the only conditioned policy to surpass the unconditional baseline.}}
\label{tab:main}
\small
\setlength{\tabcolsep}{6pt}
\begin{tabular}{@{}llccccc@{}}
\toprule
Method & Conditioning Mechanism & Score (\%) $\uparrow$ & Goal (\%) $\uparrow$ & Coll.\ (\%) $\downarrow$ & Offroad (\%) $\downarrow$ & DNF (\%) $\downarrow$ \\
\midrule
Baseline (PPO) & None (unconditional self-play) & 99.29{\scriptsize$\pm$0.08} & 99.64{\scriptsize$\pm$0.04} & \textbf{1.92}{\scriptsize$\pm$0.19} & \textbf{0.87}{\scriptsize$\pm$0.14} & 0.24{\scriptsize$\pm$0.01} \\
\midrule
\textbf{UncArms (Ours)} & \textbf{Scalar role signal ($\Delta$unc arms-race)} & \textbf{99.41}{\scriptsize$\pm$0.08}$^\dagger$ & \textbf{99.72}{\scriptsize$\pm$0.02}$^\dagger$ & 2.25{\scriptsize$\pm$0.17} & 1.13{\scriptsize$\pm$0.09} & \textbf{0.20}{\scriptsize$\pm$0.02} \\
\midrule
GIGAFLOW$^\ast$ & Task-weight vector (3D: goal, coll, off) & 94.70{\scriptsize$\pm$0.70}$^\dagger$ & 96.60{\scriptsize$\pm$0.50}$^\dagger$ & 4.60{\scriptsize$\pm$0.40} & 4.30{\scriptsize$\pm$0.70} & 1.81{\scriptsize$\pm$0.31} \\
DIAYN & Discrete skill index ($z \in \{1..5\}$) & 98.36{\scriptsize$\pm$0.18}$^\dagger$ & 98.89{\scriptsize$\pm$0.08}$^\dagger$ & 4.65{\scriptsize$\pm$0.16} & 2.29{\scriptsize$\pm$0.19} & 0.72{\scriptsize$\pm$0.01} \\
Ent-PPO & Target entropy level & 98.33{\scriptsize$\pm$0.25}$^\dagger$ & 98.98{\scriptsize$\pm$0.14}$^\dagger$ & 5.12{\scriptsize$\pm$0.36} & 2.99{\scriptsize$\pm$0.32} & 0.62{\scriptsize$\pm$0.09} \\
\bottomrule
\end{tabular}
\begin{flushleft}
\footnotesize $^\ast$GIGAFLOW-style reproduction: same architecture with concatenation-based conditioning on the 3 task reward weights ($w_\mathrm{goal}, w_\mathrm{coll}, w_\mathrm{off}$)~\cite{cusumano2025robust}. The 4.59\% score gap vs.\ Baseline demonstrates the \emph{diversity tax}, conditioning on variable task weights forces the policy to represent incompatible objectives, degrading driving competence.
\end{flushleft}
\vspace{-0.2cm}
\end{table*}

Tab.~\ref{tab:main} presents the main results. \textit{UncArms surpasses the unconditional baseline} with 99.41\% vs.\ 99.29\% goal score ($p < 0.05$), the only conditioned method in our evaluation to beat a strong self-play baseline. UncArms also achieves the lowest DNF rate (0.20\%), indicating that exposure to probing agents during training improves robustness to difficult scenarios that would otherwise cause agents to fail. We find that \emph{score}, which requires both goal achievement and collision avoidance in the same episode, is the more informative metric than raw collision rate alone. An overly conservative agent can suppress collisions by slowing or avoiding contested paths, but this also prevents goal completion and inflates the DNF rate. Such behavior is not useful driving. UncArms achieves the highest score on WOMD (99.41\% vs.\ 99.29\%), nuplan Val14 (99.12\% vs.\ 97.97\%), and Argoverse~2 (82.39\% vs.\ 78.03\%), demonstrating that it reaches goals more reliably \emph{and} collision-free across all three benchmarks. The slightly higher raw collision rate on WOMD (2.25\% vs.\ 1.92\%) is offset by more successful goal-directed behavior: of the agents that do collide, UncArms collides \emph{less} on episodes where the goal is also reached (0.29\% goal-collision overlap vs.\ 0.33\% for baseline), so the extra collisions come from active failed attempts rather than reckless driving. Among diversity-oriented baselines, UncArms exceeds DIAYN (98.36\%), entropy-conditioned PPO (98.33\%), and the GIGAFLOW-style full-vector conditioning (94.70\%).

\subsection{Zero-Shot Transfer}

\subsubsection{nuPlan}

\begin{table}[t]
\centering
\caption{Zero-shot transfer to nuplan Val14 (no fine-tuning). SDC-only: ego agent controlled by policy, background agents replay logged trajectories. \textit{UncArms achieves best transfer} on all metrics.}
\label{tab:nuplan}
\small
\setlength{\tabcolsep}{3pt}
\resizebox{\columnwidth}{!}{%
\begin{tabular}{@{}lccccc@{}}
\toprule
Method & Score (\%) $\uparrow$ & Goal (\%) $\uparrow$ & Coll.\ (\%) $\downarrow$ & Offroad (\%) $\downarrow$ & DNF (\%) $\downarrow$ \\
\midrule
\textbf{UncArms (Ours)} & \textbf{99.12}{\scriptsize$\pm$0.32} & \textbf{99.24}{\scriptsize$\pm$0.30} & \textbf{9.53}{\scriptsize$\pm$0.14} & \textbf{1.15}{\scriptsize$\pm$0.10} & \textbf{0.44}{\scriptsize$\pm$0.18} \\
Baseline (PPO) & 97.97{\scriptsize$\pm$0.57} & 98.27{\scriptsize$\pm$0.58} & 9.61{\scriptsize$\pm$0.31} & 1.36{\scriptsize$\pm$0.26} & 1.27{\scriptsize$\pm$0.45} \\
% UncControl row removed — controllability experiment excluded from revision
\midrule
DIAYN & 97.16{\scriptsize$\pm$0.75} & 97.35{\scriptsize$\pm$0.62} & 13.12{\scriptsize$\pm$0.50} & 2.31{\scriptsize$\pm$0.46} & 1.08{\scriptsize$\pm$0.23} \\
Ent-PPO & 95.67{\scriptsize$\pm$0.25} & 95.98{\scriptsize$\pm$0.39} & 12.40{\scriptsize$\pm$0.54} & 2.48{\scriptsize$\pm$0.37} & 2.47{\scriptsize$\pm$0.49} \\
GIGAFLOW$^\ast$ & 91.82{\scriptsize$\pm$1.91} & 92.02{\scriptsize$\pm$1.82} & 10.95{\scriptsize$\pm$0.66} & 2.31{\scriptsize$\pm$0.53} & 3.97{\scriptsize$\pm$1.33} \\
\bottomrule
\end{tabular}}
\end{table}

Tab.~\ref{tab:nuplan} presents the zero-shot transfer results, which provide the strongest evidence for UncArms's value. \textit{UncArms achieves 99.12\% score, surpassing the baseline by 1.15 percentage points}, with higher goal rate (99.24\% vs.\ 98.27\%), lower collision rate (9.53\% vs.\ 9.61\%), and lower offroad rate (1.15\% vs.\ 1.36\%). Note that collision rates in SDC-only mode are elevated across all methods because background agents follow fixed logged trajectories without reacting to the ego agent; the relative ordering across methods remains meaningful. 
This result parallels findings in adversarial training for image classification~\cite{madry2018towards} and sim-to-real transfer~\cite{tobin2017domain}. Models exposed to a broader distribution of challenging conditions during training develop more robust internal representations that transfer better to novel domains. In our setting, the diversity-driven curriculum forces agents to handle a wider distribution of interactive situations during training, producing a policy that is less brittle when encountering the different traffic patterns, map formats, and lane geometries of nuplan Val14. 
Diversity-oriented baselines transfer substantially worse. GIGAFLOW scores 91.82\% ($-$6.15\% vs.\ baseline) and Ent-PPO 95.67\% ($-$2.30\%). DIAYN transfers at 97.16\%, still below baseline. UncArms is the only method that consistently outperforms the baseline across all five metrics.

\textbf{Comparison of diversity mechanisms.} The entropy baseline diversifies behavior without direction as it encourages exploration broadly but does not concentrate pressure on any particular region of the policy's distribution. DIAYN diversifies the ego policy's own behavioral range but optimizes for discriminability rather than safety-relevance. Our uncertainty-seeking approach performs better as it focuses probing pressure on interactions where the policy has low confidence, potentially generating scenario diversity more aligned with policy gaps.

\subsubsection{Argoverse 2}

\begin{table}[t]
\centering
\caption{Zero-shot transfer to Argoverse 2 validation (24,988 scenarios, no fine-tuning). SDC-only control. \textit{UncArms leads on score and goal rate.}}
\label{tab:av2}
\small
\setlength{\tabcolsep}{3pt}
\resizebox{\columnwidth}{!}{%
\begin{tabular}{@{}lccccc@{}}
\toprule
Method & Score (\%) $\uparrow$ & Goal (\%) $\uparrow$ & Coll.\ (\%) $\downarrow$ & Offroad (\%) $\downarrow$ & DNF (\%) $\downarrow$ \\
\midrule
\textbf{UncArms (Ours)} & \textbf{82.39}{\scriptsize$\pm$0.68} & \textbf{93.05}{\scriptsize$\pm$0.49} & 9.84{\scriptsize$\pm$0.13} & \textbf{26.67}{\scriptsize$\pm$1.56} & \textbf{0.69}{\scriptsize$\pm$0.08} \\
Baseline (PPO) & 78.03{\scriptsize$\pm$2.79} & 89.33{\scriptsize$\pm$2.78} & \textbf{9.66}{\scriptsize$\pm$0.46} & 30.07{\scriptsize$\pm$2.61} & 1.25{\scriptsize$\pm$0.17} \\
\midrule
DIAYN & 75.00{\scriptsize$\pm$2.88} & 82.77{\scriptsize$\pm$2.90} & 12.02{\scriptsize$\pm$0.83} & 28.28{\scriptsize$\pm$2.93} & 3.09{\scriptsize$\pm$0.09} \\
Ent-PPO & 76.92{\scriptsize$\pm$0.93} & 85.18{\scriptsize$\pm$0.91} & 12.82{\scriptsize$\pm$0.72} & 27.70{\scriptsize$\pm$0.79} & 3.26{\scriptsize$\pm$0.99} \\
GIGAFLOW$^\ast$ & 69.44{\scriptsize$\pm$2.30} & 76.71{\scriptsize$\pm$1.74} & 11.21{\scriptsize$\pm$0.43} & 30.18{\scriptsize$\pm$2.09} & 4.51{\scriptsize$\pm$0.81} \\
\bottomrule
\end{tabular}}
\vspace{-0.2cm}
\end{table}

Tab.~\ref{tab:av2} presents zero-shot transfer results on Argoverse~2~\cite{wilson2023argoverse2generativedatasets}. AV2 lacks explicit road-edge annotations so we approximate edges via the drivable-area polygon boundary, which cuts diagonally across open intersections and flags legal turns as off-road. \textit{Offroad rates are therefore inflated for all methods} (26–30\% vs.\ $<$2\% on nuPlan). Score and goal rate remain reliable comparators. 
Despite this limitation, the relative ordering is consistent with nuPlan where \textit{UncArms achieves the highest score (82.39\%) and goal rate (93.05\%)}, outperforming the baseline by $+$4.36\% and $+$3.72\% respectively. The margin is larger here than on nuPlan ($+$1.15\%), suggesting that the diversity-driven curriculum confers additional benefit on more geometrically diverse scenarios. Diversity baselines again transfer poorly. GIGAFLOW scores 69.44\% ($-$8.59\% vs.\ baseline), DIAYN 75.00\% ($-$3.03\%), and Ent-PPO 76.92\% ($-$1.11\%). UncArms is the only method to outperform the unconditional baseline on all three transfer datasets (WOMD, nuPlan, and AV2).

\subsection{Uncertainty as Difficulty Detector}

\begin{figure}[t]
\centering
\includegraphics[width=0.8\columnwidth]{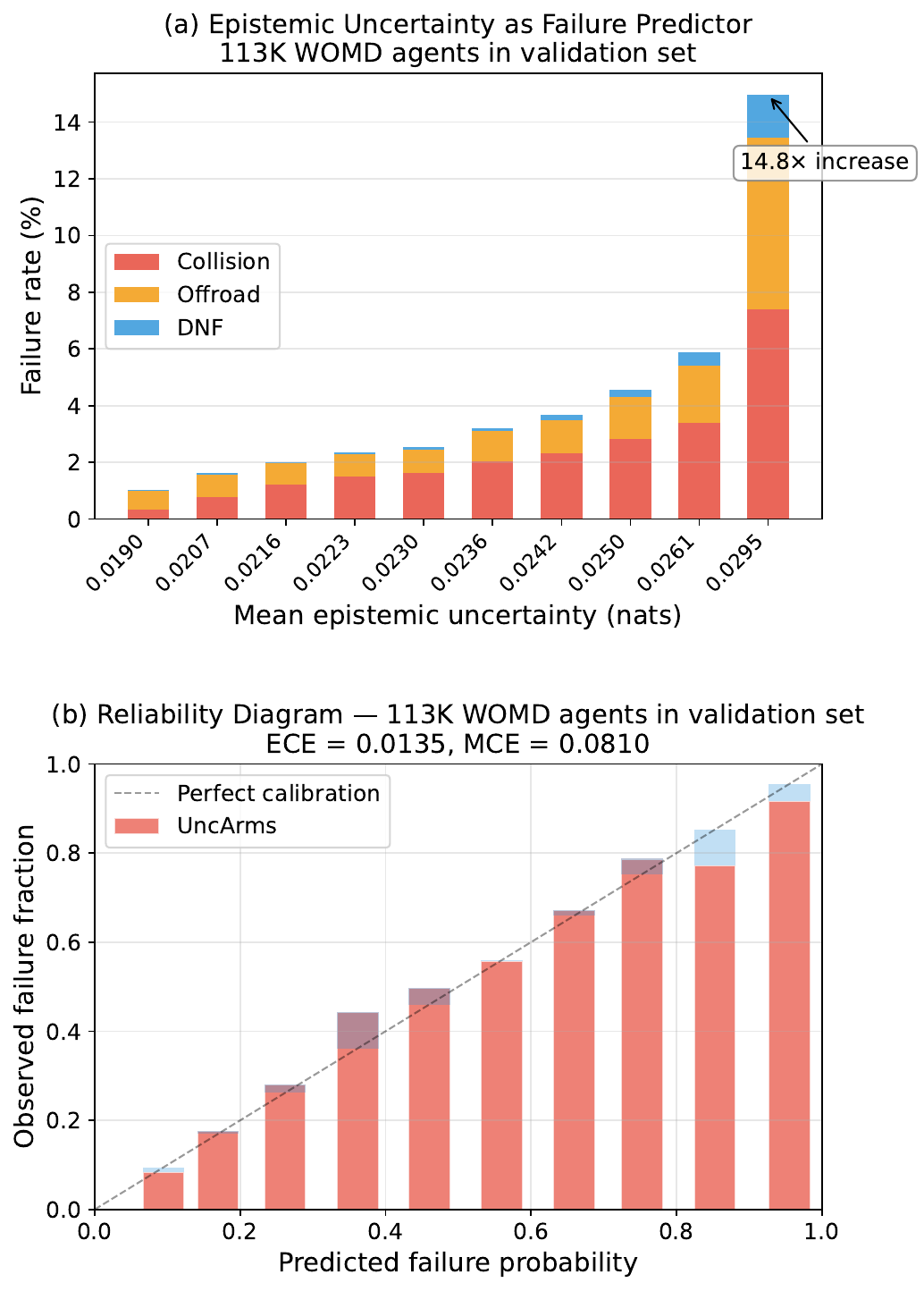}
\caption{Epistemic uncertainty as failure predictor and calibration analysis (UncArms, 113K agents across 10K WOMD validation maps, $w_\mathrm{unc}{=}0$). \textbf{(a)}~Combined failure rate (collision + offroad + DNF) by uncertainty decile. \textbf{(b)}~Reliability diagram: logistic mapping fitted on training data, evaluated on held-out validation.}
\label{fig:calibration}
\vspace{-0.2cm}
\end{figure}

Fig.~\ref{fig:calibration}(a) demonstrates that epistemic uncertainty is a strong predictor of driving failure. Binning 113K agent-level episodes by MC-Dropout uncertainty decile reveals a monotonic 14.8$\times$ increase in combined failure rate (collision + offroad + DNF) from the lowest to highest decile (1.01\% vs.\ 14.96\%). All three failure modes scale with uncertainty, with collisions dominating and DNF exhibiting the steepest relative growth. Fig.~\ref{fig:calibration}(b) shows a reliability diagram: a logistic mapping from uncertainty to failure probability, fitted on a random 1000-map subsample of the training set, is evaluated on 10K held-out validation maps (113K episodes). The resulting Expected Calibration Error (ECE$=$0.014) confirms that the uncertainty signal is not merely a good ranker but also yields well-calibrated failure probabilities after a simple post-hoc transformation. 
This finding motivates the self-play mechanism: if uncertainty reliably indicates difficulty, then rewarding agents for \emph{increasing} neighbors' uncertainty naturally generates challenging interactions. Beyond its role in training, this difficulty signal is a practical deployment tool as a single UncArms checkpoint simultaneously provides best-in-class driving performance (Tab.~\ref{tab:main}) and a calibrated per-scenario failure predictor (Fig.~\ref{fig:calibration}) without requiring ground-truth labels or separate evaluation runs.

\subsection{Behavioral Diversity via LSTM Hidden States}

\begin{table}[t]
\centering
\caption{LSTM hidden-state diversity.}
\label{tab:lstm}
\small
\begin{tabular}{@{}lccc@{}}
\toprule
Method & Coverage $\uparrow$ & Eff.\ Dim.\ $\uparrow$ & Spread \\
\midrule
\textbf{UncArms (Ours)} & \textbf{0.357} & \textbf{110} & \textbf{34.33} \\
% UncControl row removed — controllability experiment excluded from revision
Baseline (PPO) & 0.193 & 94 & 33.25 \\
\bottomrule
\end{tabular}
\end{table}

\begin{table}[t]
\centering
\caption{Ablation isolating the adversarial training contribution (10K WOMD maps, 4 seeds). MC-Dropout alone affects the performance while the arms-race mechanism more than compensates.}
\label{tab:ablation}
\small
\setlength{\tabcolsep}{3pt}
\begin{tabular}{@{}lcc@{}}
\toprule
Method & Arms-Race & Score (\%) \\
\midrule
Baseline (PPO, no dropout) & \texttimes & 99.29{\scriptsize$\pm$0.08} \\
Baseline + MC-Dropout & \texttimes & 98.50{\scriptsize$\pm$0.08} \\
\textbf{UncArms (Ours)} & \checkmark & \textbf{99.41}{\scriptsize$\pm$0.08} \\
\bottomrule
\end{tabular}
\vspace{-0.2cm}
\end{table}

To quantify whether adversarial training yields genuinely different behaviors or merely different action magnitudes, we analyze the LSTM hidden states during evaluation (Tab.~\ref{tab:lstm}). The LSTM encodes the policy's internal ``situation model''. Distinct hidden states correspond to distinct behavioral modes the policy has learned to represent.

\textbf{Coverage} measures what fraction of the representational space the policy actually uses. We project hidden states onto 3 principal components, divide the resulting space into a $20^3{=}8000$-cell grid, and count the fraction of cells visited. UncArms occupies 35.7\% of the grid vs.\ 19.3\% for the Baseline, an 85\% increase, meaning the adversarial policy activates a far broader set of behavioral modes during evaluation rather than repeatedly visiting the same few states.

\textbf{Effective dimensionality} counts how many independent directions the hidden states spread across (PCA components needed for 95\% of variance). UncArms uses 110 dimensions vs.\ 94 for the Baseline, indicating more uniform representation spread rather than collapse along few dominant axes.

\textbf{Spread} (trace of the covariance matrix) measures total variance. The Baseline and UncArms have similar spread (33.25 vs.\ 34.33), but UncArms distributes this variance more uniformly across space (higher coverage) rather than concentrating in dense clusters with empty gaps.

Together, these metrics indicate that the diversity-driven curriculum forces the policy to develop a richer internal set of behavioral modes, directly explaining its superior zero-shot transfer: a policy exposed to more situations during training is better equipped for novel scenarios.

\subsection{Ablation Study}

\subsubsection{MC-Dropout vs.\ Adversarial Training}

A natural question is whether UncArms's gains stem from its adversarial training mechanism or simply from the regularization effect of MC-Dropout. Tab.~\ref{tab:ablation} isolates this factor. Adding MC-Dropout to the baseline \emph{without} adversarial training \emph{reduces} the score by 0.79\%. UncArms uses the same architecture but surpasses the clean baseline by 0.12\%, a net swing of +0.91\% attributable entirely to the arms-race mechanism. MC-Dropout is a necessary enabler (providing the uncertainty signal for $\Delta$unc rewards) but is not itself beneficial. The gains come from the adversarial curriculum it enables.

\subsubsection{Neighborhood Size and Warmup Schedule}

\begin{table}[t]
\centering
\caption{Neighborhood size ($K$) and warmup ablations.}
\label{tab:ablation_k_warmup}
\small
\setlength{\tabcolsep}{4pt}
\begin{tabular}{@{}lc@{}}
\toprule
\textit{(a) Neighborhood size $K$}  & Score (\%) $\uparrow$\\
\midrule
$K=1$ & 98.97{\scriptsize$\pm$0.24} \\
\textbf{$K=3$ (Ours)} & \textbf{99.41}{\scriptsize$\pm$0.08} \\
$K=5$ & 99.22{\scriptsize$\pm$0.09} \\
$K=10$ & 98.99{\scriptsize$\pm$0.20} \\
\midrule
\textit{(b) Warmup schedule}  & Score (\%) $\uparrow$ \\
\midrule
\textbf{With warmup (Ours)} & \textbf{99.41}{\scriptsize$\pm$0.08} \\
No warmup & 98.84{\scriptsize$\pm$0.24} \\
\bottomrule
\end{tabular}
\end{table}

\begin{table}[t]
\centering
\caption{Effect of conditioning mechanism on WOMD score. All variants use the same arms-race training. Only the role-signal modulation differs.}
\label{tab:ablation_conditioning}
\small
\setlength{\tabcolsep}{4pt}
\begin{tabular}{@{}lcc@{}}
\toprule
Method & Score (\%) $\uparrow$ \\
\midrule
Baseline (PPO)  & 99.29{\scriptsize$\pm$0.08} \\
\midrule
UncArms (no role assignment) & 99.18{\scriptsize$\pm$0.30} \\
UncArms (Concatentation) & 99.35{\scriptsize$\pm$0.07} \\
\textbf{UncArms (FiLM, Ours)} & \textbf{99.41}{\scriptsize$\pm$0.08} \\
\bottomrule
\end{tabular}
\vspace{-0.2cm}
\end{table}

Tab.~\ref{tab:ablation_k_warmup}(a) ablates the neighborhood size $K$ used in the $\Delta$unc reward. $K{=}3$ achieves the best score. Both extremes ($K{=}1$ and $K{=}10$) degrade performance. With $K{=}1$, the probing signal is too sparse to generate a consistent curriculum. With $K{=}10$, averaging over distant agents dilutes the signal from the most proximate and relevant interactions. $K{=}3$ balances local interaction density with signal stability. Tab.~\ref{tab:ablation_k_warmup}(b) shows that removing the 750M-step driving-only warmup degrades score by 0.57\%. The warmup first establishes a competent base policy before probing pressure is introduced where probing agents with an existing driving skill can navigate toward uncertain neighbors and generate meaningful interactions, yielding a richer uncertainty-driven curriculum that improves the final neutral policy.

\subsubsection{Conditioning Mechanism}

Tab.~\ref{tab:ablation_conditioning} isolates the role of the conditioning mechanism. Running the arms-race without any role assignment falls \emph{below} the unconditional baseline ($-$0.11\%), confirming that the probing reward alone is insufficient as agents cannot adopt distinct exploratory strategies if the policy cannot perceive its role. Concatenation conditioning recovers and surpasses the baseline ($+$0.06\%), demonstrating that role awareness itself is the key contributor. FiLM's multiplicative modulation (Ours) adds a further $+$0.06\%, suggesting feature-wise gating provides a small but consistent benefit over additive conditioning.

\subsubsection{Probing Fraction}

\begin{table}[t]
\centering
\caption{Effect of probing fraction of UncArm on WOMD score.}
\label{tab:ablation_probing_frac}
\small
\setlength{\tabcolsep}{4pt}
\begin{tabular}{@{}lccc@{}}
\toprule
Neutral (\%) & Probing (\%) & Score (\%) $\uparrow$ \\
\midrule
25 & 75 & 99.04{\scriptsize$\pm$0.53} \\
 \textbf{50} & \textbf{50} & \textbf{99.41}{\scriptsize$\pm$0.08} \\
75 & 25 & 99.40{\scriptsize$\pm$0.03} \\
\bottomrule
\end{tabular}
\vspace{-0.2cm}
\end{table}

\begin{figure}[t]
    \centering
    \includegraphics[width=0.8\columnwidth]{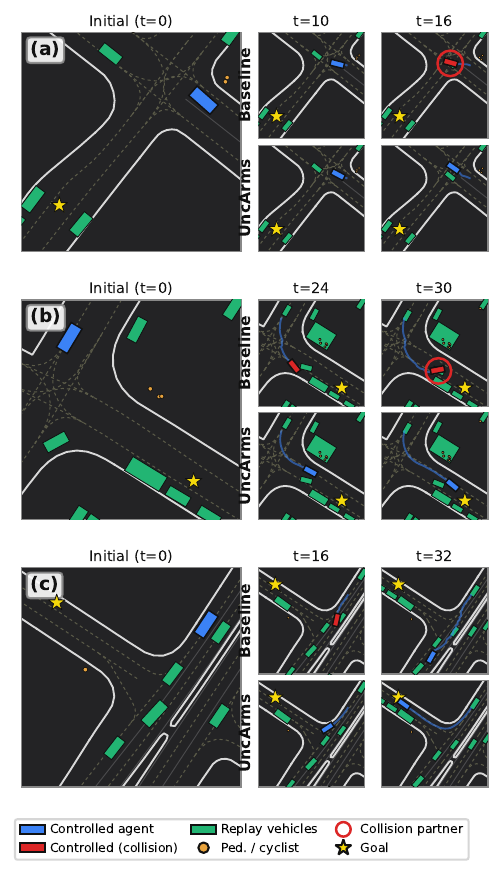}
    \caption{Qualitative safety comparison under neutral conditioning ($w_\mathrm{unc}{=}0$). For each scenario, the left panel shows the initial state, and the $2{\times}2$ matrix shows the controlled ego (blue, red on collision) for \textbf{Baseline} (top) and \textbf{UncArms} (bottom) at two timesteps. Other agents replay their logged trajectories (green vehicles, amber pedestrians/cyclists), the yellow star marks the ego goal, and the red ring marks the collision partner.}
    \label{fig:qualitative_safety}
    \vspace{-0.2cm}
\end{figure}

Tab.~\ref{tab:ablation_probing_frac} varies the fraction of neutral agents per episode. The 50/50 split (UncArms) and the 75\% neutral split (E37) perform nearly identically (99.41\% vs.\ 99.40\%), indicating that a minority of probing agents is sufficient to generate a useful curriculum. Reducing neutral agents to 25\% (E36) degrades performance substantially ($-$0.37\% vs.\ UncArms, variance $\pm$0.53), because too few neutral agents weaken the task-reward anchor and destabilize training. The asymmetry suggests that curriculum quality is limited by the diversity of probing behaviors, not by their quantity. Once enough probing pressure exists to surface novel scenarios, additional probers do not further help.

\subsection{Qualitative Results}

Fig.~\ref{fig:qualitative_safety} contrasts the unconditional Baseline with UncArms on three representative WOMD scenarios, both evaluated under identical neutral conditioning ($w_\mathrm{unc}{=}0$), so the differences stem from adversarial training rather than any test-time signal. In~(a), UncArms shows greater negotiation capacity in a dense crossing conflict where it yields and waits for the other agents to pass first, then proceeds through the conflict zone and reaches its goal, whereas the Baseline drives into the conflict and collides. In~(b), UncArms avoids a crossing-vehicle collision that the Baseline incurs, and in~(c) it completes a route the Baseline fails to finish. These safer, negotiation-aware behaviors emerge purely from the adversarial arms-race during training.

\begin{figure}[t]
    \centering
    \includegraphics[width=0.8\columnwidth]{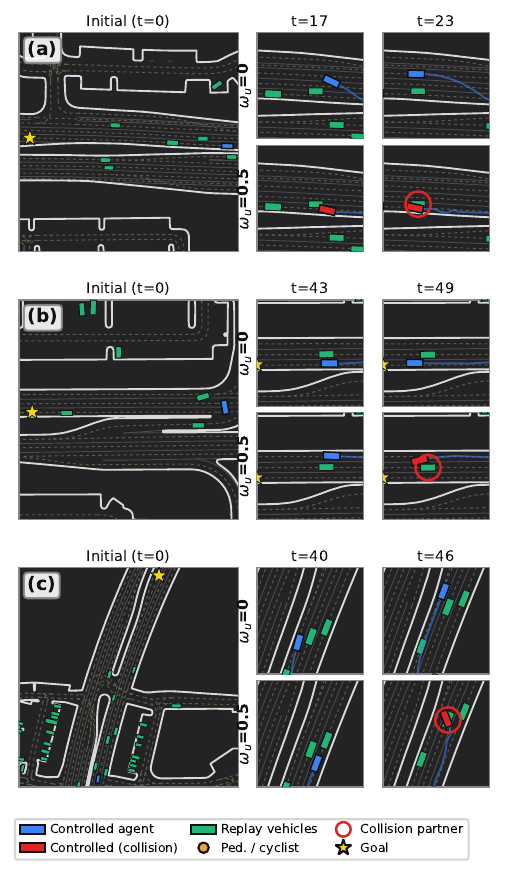}
    \caption{Role-conditioned behavior from a single UncArms checkpoint on three WOMD scenarios. Left: initial scene with the goal visible. $2{\times}2$ matrix: ego at two timesteps under \textbf{neutral} conditioning ($w_\mathrm{unc}{=}0$, top) and \textbf{probing} conditioning ($w_\mathrm{unc}{=}0.5$, bottom). Blue: controlled ego. red: ego at collision. green: replay vehicles. amber: pedestrians/cyclists. red ring: collision partner.}
    \label{fig:qualitative_roles}
    \vspace{-0.2cm}
\end{figure}

Fig.~\ref{fig:qualitative_roles} illustrates how a single UncArms checkpoint produces qualitatively distinct behaviors depending solely on the conditioning signal $w_\mathrm{unc}$ at inference time. Under neutral conditioning ($w_\mathrm{unc}{=}0$), the ego pursues a conservative, goal-directed strategy and avoids collisions in all three scenarios. Switching to probing conditioning ($w_\mathrm{unc}{=}0.5$) from the \emph{same checkpoint} produces a novelty-seeking agent that actively explores uncertain, high-interaction regions of the scene. Collisions arise as a natural byproduct of this exploratory drive rather than as an explicit objective. This role flexibility emerges at no additional cost as the checkpoint is shared and only the conditioning signal changes at test time. A current limitation of the neutral role is that, while it achieves higher scores, its trajectories do not necessarily comply with traffic regulations such as lane-keeping or right-of-way rules, since these constraints are absent from the reward definition. Addressing explicit rule compliance is left for future work.

%\section{Limitations and Future Work}

%\textbf{Discrete action space.} Our policy uses a discretized 7$\times$13 action grid. While sufficient for highway and urban driving at the current resolution, continuous actions with distributional critics may enable more nuanced probing behaviors.

%\textbf{MC-Dropout computational cost.} The $K{=}5$ forward passes per step during training rollouts add computational cost to the training loop; evaluation requires no additional passes as uncertainty is not used at test time. More efficient uncertainty methods such as evidential deep learning~\cite{sensoy2018evidentialdeeplearningquantify} or single-pass deterministic uncertainty~\cite{vanamersfoort2020uncertaintyestimationusingsingle} could reduce training cost while preserving the self-play mechanism.

%\textbf{Self-play stability.} The uncertainty-driven self-play requires careful warmup scheduling (750M steps of driving-only before probing pressure is introduced). Adaptive probing magnitude based on neutral policy improvement rate could make training more robust to hyperparameter choices.

%\textbf{Scalability of the approach.} Our ablation shows $K{=}3$ nearest neighbors is optimal (Table~\ref{tab:ablation_k_warmup}), but investigating adversary fraction and multi-hop uncertainty propagation could yield further improvements.

\section{Conclusion}

We presented UncArms, an uncertainty-driven self-play scheme for multi-agent driving that uses changes in epistemic uncertainty as the probing reward. In the proposed mechanism, the reward probing agents increase neighbors' confusion and let neutral agents adapt. This yields self-play dynamics that expose the neutral policy to a broader distribution of challenging interactions. Among the conditioned methods we evaluate, UncArms is the only one to outperform the unconditional self-play baseline with 99.41\% vs.\ 99.29\% on WOMD, 99.12\% vs.\ 97.97\% on zero-shot nuplan Val14 transfer, and 82.39\% vs.\ 78.03\% on Argoverse~2, with no fine-tuning. The 85\% increase in LSTM hidden-state coverage explains the transfer advantage since the diversity-driven curriculum forces the policy to internalize broader behaviors.

The core insight is that epistemic uncertainty, measuring where the policy has the lowest confidence, provides a cheap, automatic proxy for behavioral diversity in multi-agent settings. Rather than hand-engineering diverse training scenarios, probing agents find the hard cases automatically by targeting precisely the situations where the current policy is least confident. Additionally, epistemic uncertainty serves as a calibrated difficulty detector (14.8$\times$ failure rate in highest vs.\ lowest decile, ECE$=$0.014), providing a signal for prioritizing challenging scenarios without ground-truth labels.

\section*{ACKNOWLEDGMENT}

This work is a result of the joint research project STADT:up (19A22006O). The project is supported by the German Federal Ministry for Economic Affairs and Energy (BMWE), based on a decision of the German Bundestag. The author is solely responsible for the content of this publication.

{\footnotesize
\bibliographystyle{IEEEtran}
\bibliography{references}

@String { CVPR                 = {IEEE/CVF Conf.~on Computer Vision and Pattern Recognition} }

@String { ICRA                 = {IEEE Int.~Conf.~on Robotics and Automation} }

@String { IROS                 = {Int.~Conf.~on Intelligent Robots and Systems} }

@String { ROBOTICS             = {Proc.~of the Int.~Conf.~on Robotics: Science and Systems} }

@article{gal2016dropoutbayesianapproximationrepresenting,
      title={Dropout as a Bayesian Approximation: Representing Model Uncertainty in Deep Learning}, 
      author={Yarin Gal and Zoubin Ghahramani},
      year={2016},
  journal={arXiv preprint arXiv:1506.02142},
}

@article{lakshminarayanan2017simple,
      title={Simple and Scalable Predictive Uncertainty Estimation using Deep Ensembles}, 
      author={Balaji Lakshminarayanan and Alexander Pritzel and Charles Blundell},
      year={2017},
  journal={arXiv preprint arXiv:1612.01474},
}

@article{perez2018film,
      title={FiLM: Visual Reasoning with a General Conditioning Layer}, 
      author={Ethan Perez and Florian Strub and Harm de Vries and Vincent Dumoulin and Aaron Courville},
      year={2017},
  journal={arXiv preprint arXiv:1709.07871},
}

@article{pathak2017curiositydrivenexplorationselfsupervisedprediction,
      title={Curiosity-driven Exploration by Self-supervised Prediction}, 
      author={Deepak Pathak and Pulkit Agrawal and Alexei A. Efros and Trevor Darrell},
      year={2017},
  journal={arXiv preprint arXiv:1705.05363},
}

@article{burda2018largescalestudycuriositydrivenlearning,
      title={Large-Scale Study of Curiosity-Driven Learning}, 
      author={Yuri Burda and Harri Edwards and Deepak Pathak and Amos Storkey and Trevor Darrell and Alexei A. Efros},
      year={2018},
  journal={arXiv preprint arXiv:1808.04355},
}

@article{caesar2020nuscenesmultimodaldatasetautonomous,
      title={nuScenes: A multimodal dataset for autonomous driving}, 
      author={Holger Caesar and Varun Bankiti and Alex H. Lang and others},
      year={2020},
  journal={arXiv preprint arXiv:1903.11027},
}

@inproceedings{wang2021advsim,
  title={Advsim: Generating safety-critical scenarios for self-driving vehicles},
  author={Wang, Jingkang and Pun, Ava and Tu, James and Manivasagam, Sivabalan and Sadat, Abbas and Casas, Sergio and Ren, Mengye and Urtasun, Raquel},
  booktitle=CVPR,
  pages={9909--9918},
  year={2021}
}

@article{hanselmann2022kinggeneratingsafetycriticaldriving,
      title={KING: Generating Safety-Critical Driving Scenarios for Robust Imitation via Kinematics Gradients}, 
      author={Niklas Hanselmann and Katrin Renz and Kashyap Chitta and Apratim Bhattacharyya and Andreas Geiger},
      year={2022},
  journal={arXiv preprint arXiv:2204.13683},
}

@article{nuplan,
  title={nuplan: A closed-loop ml-based planning benchmark for autonomous vehicles},
  author={Caesar, Holger and Kabzan, Juraj and Tan, Kok Seang and Fong, Whye Kit and Wolff, Eric and Lang, Alex and Fletcher, Luke and Beijbom, Oscar and Omari, Sammy},
  journal={arXiv preprint arXiv:2106.11810},
  year={2021}
}

@article{vinyals2019grandmaster,
  title={Grandmaster level in StarCraft II using multi-agent reinforcement learning},
  author={Vinyals, Oriol and Babuschkin, Igor and Czarnecki, Wojciech M and others},
  journal={nature},
  volume={575},
  number={7782},
  pages={350--354},
  year={2019},
}

@article{suarez2024pufferlibmakingreinforcementlearning,
      title={PufferLib: Making Reinforcement Learning Libraries and Environments Play Nice}, 
      author={Joseph Suarez},
      year={2024},
  journal={arXiv preprint arXiv:2406.12905},
}

@article{ettinger2021largescaleinteractivemotion,
      title={Large Scale Interactive Motion Forecasting for Autonomous Driving : The Waymo Open Motion Dataset}, 
      author={Scott Ettinger and Shuyang Cheng and Benjamin Caine and others},
      year={2021},
  journal={arXiv preprint arXiv:2104.10133},
}

@article{schulman2017proximalpolicyoptimizationalgorithms,
      title={Proximal Policy Optimization Algorithms}, 
      author={John Schulman and Filip Wolski and Prafulla Dhariwal and others},
      year={2017},
  journal={arXiv preprint arXiv:1707.06347},
}

@article{cornelisse2025building,
  title={Building reliable sim driving agents by scaling self-play},
  author={Cornelisse, Daphne and Pandya, Aarav and Joseph, Kevin and Su{\'a}rez, Joseph and Vinitsky, Eugene},
  journal={arXiv preprint arXiv:2502.14706},
  year={2025}
}

@article{cusumano2025robust,
  title={Robust autonomy emerges from self-play},
  author={Cusumano-Towner, Marco and Hafner, David and others},
  journal={arXiv preprint arXiv:2502.03349},
  year={2025}
}

@article{marvi2025evidential,
  title={Evidential Uncertainty Estimation for Multi-Modal Trajectory Prediction},
  author={Marvi, Sajad and Rist, Christoph and Schmidt, Julian and Jordan, Julian and Valada, Abhinav},
  journal={arXiv preprint arXiv:2503.05274},
  year={2025}
}

@inproceedings{schmalstieg2022learning,
  title={Learning long-horizon robot exploration strategies for multi-object search in continuous action spaces},
  author={Schmalstieg, Fabian and Honerkamp, Daniel and Welschehold, Tim and Valada, Abhinav},
  booktitle={The Int. Symposium of Robotics Research},
  pages={52--66},
  year={2022},
}

@inproceedings{eysenbach2019diayn,
  title={Diversity is All You Need: Learning Skills without a Reward Function},
  author={Eysenbach, Benjamin and Gupta, Abhishek and Ibarz, Julian and Levine, Sergey},
  booktitle={Int. Conf. on Learning Representations},
  year={2019}
}

@inproceedings{haarnoja2018sac,
  title={Soft Actor-Critic: Off-Policy Maximum Entropy Deep Reinforcement Learning with a Stochastic Actor},
  author={Haarnoja, Tuomas and Zhou, Aurick and Abbeel, Pieter and Levine, Sergey},
  booktitle={Int. Conf. on Machine Learning},
  pages={1861--1870},
  year={2018}
}

@inproceedings{rowe2024ctrlsim,
  title={CtRL-Sim: Reactive and Controllable Driving Agents with Offline Reinforcement Learning},
  author={Rowe, Luke and Ethier, Roger and Rajwanshi, Aditya and Zeng, Yuhan},
  booktitle={Conf. on Robot Learning},
  year={2024}
}

@inproceedings{matsgym2024,
  title={MATS-Gym: Multi-Agent Traffic Scenario Benchmark for Autonomous Driving},
  author={Scheffe, Patrick and Xu, Zikang and Kloock, Maximilian and Alrifaee, Bassam},
  booktitle={IEEE Intel. Vehicles Symposium},
  year={2024}
}

@inproceedings{madry2018towards,
  title={Towards Deep Learning Models Resistant to Adversarial Attacks},
  author={Madry, Aleksander and Makelov, Aleksandar and Schmidt, Ludwig and Tsipras, Dimitris and Vladu, Adrian},
  booktitle={Int. Conf. on Learning Representations},
  year={2018}
}

@inproceedings{tobin2017domain,
  title={Domain Randomization for Transferring Deep Neural Networks from Simulation to the Real World},
  author={Tobin, Josh and Fong, Rachel and Ray, Alex and Schneider, Jonas and Zaremba, Wojciech and Abbeel, Pieter},
  booktitle=IROS,
  year={2017}
}

@inproceedings{lutjens2019saferl,
  title={Safe Reinforcement Learning with Model Uncertainty Estimates},
  author={L{\"u}tjens, Bj{\"o}rn and Everett, Michael and How, Jonathan P},
  booktitle=ICRA,
  year={2019}
}

@article{kahn2017uncertaintyaware,
  title={Uncertainty-Aware Reinforcement Learning for Collision Avoidance},
  author={Kahn, Gregory and Villaflor, Adam and Pong, Vitchyr and Abbeel, Pieter and Levine, Sergey},
  journal={arXiv preprint arXiv:1702.01182},
  year={2017}
}

@inproceedings{dennis2020emergent,
  title={Emergent Complexity and Zero-shot Transfer via Unsupervised Environment Design},
  author={Dennis, Michael and Jaques, Natasha and Vinitsky, Eugene and Baez, Alexandre and Singh, Sanjiban and Levine, Sergey and Mordatch, Igor},
  booktitle={Adv. in Neural Information Processing Systems},
  year={2020}
}

@inproceedings{pinto2017robust,
  title={Robust Adversarial Reinforcement Learning},
  author={Pinto, Lerrel and Davidson, James and Sukthankar, Rahul and Gupta, Abhinav},
  booktitle={Int. Conf. on Machine Learning},
  year={2017}
}

@inproceedings{wilson2023argoverse2generativedatasets,
  title={Argoverse 2: Next Generation Datasets for Self-Driving Perception and Forecasting},
  author={Wilson, Benjamin and Qi, William and Aber, Tanmay and others},
  booktitle={Adv. in Neural Information Processing Systems},
  year={2023}
}

@misc{pufferdrive2025github,
  author = {Daphne Cornelisse and Spencer Cheng and Pragnay Mandavilli and others},
  title = {{PufferDrive}: A Fast and Friendly Driving Simulator for Training and Evaluating {RL} Agents},
  url = {https://github.com/Emerge-Lab/PufferDrive},
  version = {2.0.0},
  year = {2025},
}

@article{wei2026copark,
  title={CoPark: Learning Reactive Parking via Self-Play},
  author={Wei, Jiarong and Chen, Yanxing and Song, Sinuo and Wu, Yin and Rehr, Anna and Valada, Abhinav},
  journal={arXiv preprint arXiv:2606.04149},
  year={2026}
}

@article{wei2026counterplay,
  title={CounterPlay: Counterfactual Post-Training for Self-Play Driving Policies},
  author={Wei, Jiarong and Wu, Yin and He, Runkai and Valada, Abhinav},
  journal={arXiv preprint arXiv:2609.21617},
  year={2026}
}

@article{schmalstieg2023learning,
  title={Learning hierarchical interactive multi-object search for mobile manipulation},
  author={Schmalstieg, Fabian and Honerkamp, Daniel and Welschehold, Tim and Valada, Abhinav},
  journal={IEEE Robotics and Automation Letters},
  volume={8},
  number={12},
  pages={8549--8556},
  year={2023},
}

@article{schmidt2025graphpilot,
  title={Graphpilot: Grounded scene graph conditioning for language-based autonomous driving},
  author={Schmidt, Fabian and Enzweiler, Markus and Valada, Abhinav},
  journal={arXiv preprint arXiv:2511.11266},
  year={2025}
}

@article{intsim2025,
      title={Safety-Critical Traffic Simulation with Adversarial Transfer of Driving Intentions}, 
      author={Zherui Huang and Xing Gao and Guanjie Zheng and Licheng Wen and Xuemeng Yang and Xiao Sun},
      year={2025},
      journal={arXiv preprint arXiv:2503.05180},
}

@article{safer2026,
      title={SaFeR: Safety-Critical Scenario Generation for Autonomous Driving Test via Feasibility-Constrained Token Resampling}, 
      author={Jinlong Cui and Fenghua Liang and Guo Yang and Chengcheng Tang and Jianxun Cui},
      year={2026},
      journal={arXiv preprint arXiv:2603.04071},
}
}

\end{document}